%% file: main.tex
\documentclass[runningheads]{llncs}
\usepackage{makeidx}
\usepackage{graphicx}
\usepackage{amsmath,amssymb} 
\usepackage{color}

\usepackage{acronym}
\usepackage{float}
\usepackage{subfig}
\usepackage{siunitx}
\usepackage{dirtree}
\usepackage{hyperref}
\usepackage{graphicx,caption,lipsum}
\usepackage[normalem]{ulem}

\usepackage{xspace}
\makeatletter
\DeclareRobustCommand\onedot{\futurelet\@let@token\@onedot}
\def\@onedot{\ifx\@let@token.\else.\null\fi\xspace}

\def\eg{\emph{e.g}\onedot} 
\def\ie{\emph{i.e}\onedot} 
\def\cf{\emph{c.f}\onedot}

\acrodef{imu}[IMU]{inertial measurement unit}
\acrodef{fpga}[FPGA]{field-programmable gate array}
\acrodef{slam}[SLAM]{simultaneous localization and mapping}
\acrodef{vio}[VIO]{visual-inertial odometry}
\acrodef{vo}[VO]{visual odometry}
\acrodef{ad}[AD]{autonomous driving}
\acrodef{6dof}[6DoF]{six degrees of freedom}
\acrodef{rpe}[RPE]{relative pose error}
\acrodef{rms}[RMS]{root mean square}
\acrodef{sfm}[SfM]{structure from motion}
\acrodef{rtk}[RTK]{real-time kinematic}
\acrodef{gnss}[GNSS]{global navigation satellite system}
\acrodef{ins}[INS]{inertial navigation system}
\acrodef{ate}[ATE]{absolute trajectory error}

\begin{document}
	\pagestyle{headings}
	\mainmatter

	\def\GCPR20SubNumber{***}

	\title{4Seasons: A Cross-Season Dataset for Multi-Weather SLAM in Autonomous Driving}

	\titlerunning{4Seasons: A Cross-Season Dataset for Multi-Weather SLAM in AD}
	\authorrunning{Wenzel et al.}
	\author{Patrick Wenzel\inst{1,2} \and Rui Wang\inst{1,2} \and Nan Yang\inst{1,2} \and Qing Cheng\inst{2} \and Qadeer Khan\inst{1,2} \and Lukas von Stumberg\inst{1,2} \and Niclas Zeller\inst{2} \and Daniel Cremers\inst{1,2}}
	\institute{Technical University of Munich \and Artisense \\ \email{patrick.wenzel@tum.de}}

	\maketitle
	
	\setcounter{footnote}{0}

    \input{sections/abstract.tex}
    \input{sections/introduction.tex}
    \input{sections/related_work.tex}
    \input{sections/system_overview.tex}
    \input{sections/scenarios.tex}
    \input{sections/tasks.tex}
    \input{sections/conclusion.tex}

	\bibliographystyle{splncs04}
	\bibliography{073-main}

\end{document}

%% file: sections/abstract.tex
\begin{abstract}
We present a novel dataset covering seasonal and challenging perceptual conditions for autonomous driving. Among others, it enables research on visual odometry, global place recognition, and map-based re-localization tracking. The data was collected in different scenarios and under a wide variety of weather conditions and illuminations, including day and night. This resulted in more than \SI{350}{km} of recordings in nine different environments ranging from multi-level parking garage over urban (including tunnels) to countryside and highway. We provide globally consistent reference poses with up-to centimeter accuracy obtained from the fusion of direct stereo visual-inertial odometry with RTK-GNSS. The full dataset is available at~\url{https://go.vision.in.tum.de/4seasons}.
\keywords{autonomous driving, long-term localization, SLAM, visual learning, visual odometry}
\end{abstract}

%% file: sections/introduction.tex
\section{Introduction}\label{sec:introduction}

During the last decade, research on \ac{vo} and \ac{slam} has made tremendous strides~\cite{newcombe2011dtam,engel2014lsd,mur2015orb,engel2017direct} particularly in the context of \ac{ad}~\cite{engel2015stereolsd,wang2017stereoDSO,yang2018dvso,mur2017orb2}. One reason for this progress has been the publication of large-scale datasets~\cite{Cordts2016city,Geiger2013IJRR,Caesar2019nuscenes} tailored for benchmarking these methods. Naturally, the next logical step towards progressing research in the direction of visual \ac{slam} has been to make it robust under dynamically changing and challenging conditions. This includes \ac{vo}, \eg at night or rain, as well as long-term place recognition and re-localization against a pre-built map. In this regard, the advent of deep learning has exhibited itself to be a promising potential in complementing the performance of visual \ac{slam}~\cite{Dusmanu2019CVPR,jung2019corl,gn-net-2020,jaramillo2017direct}. Therefore, it has become all the more important to have datasets that are commensurate with handling the challenges of any real-world environment while also being capable of discerning the performance of state-of-the-art approaches. 

\begin{figure}[t]
    \centering
    \includegraphics[width=0.8\linewidth]{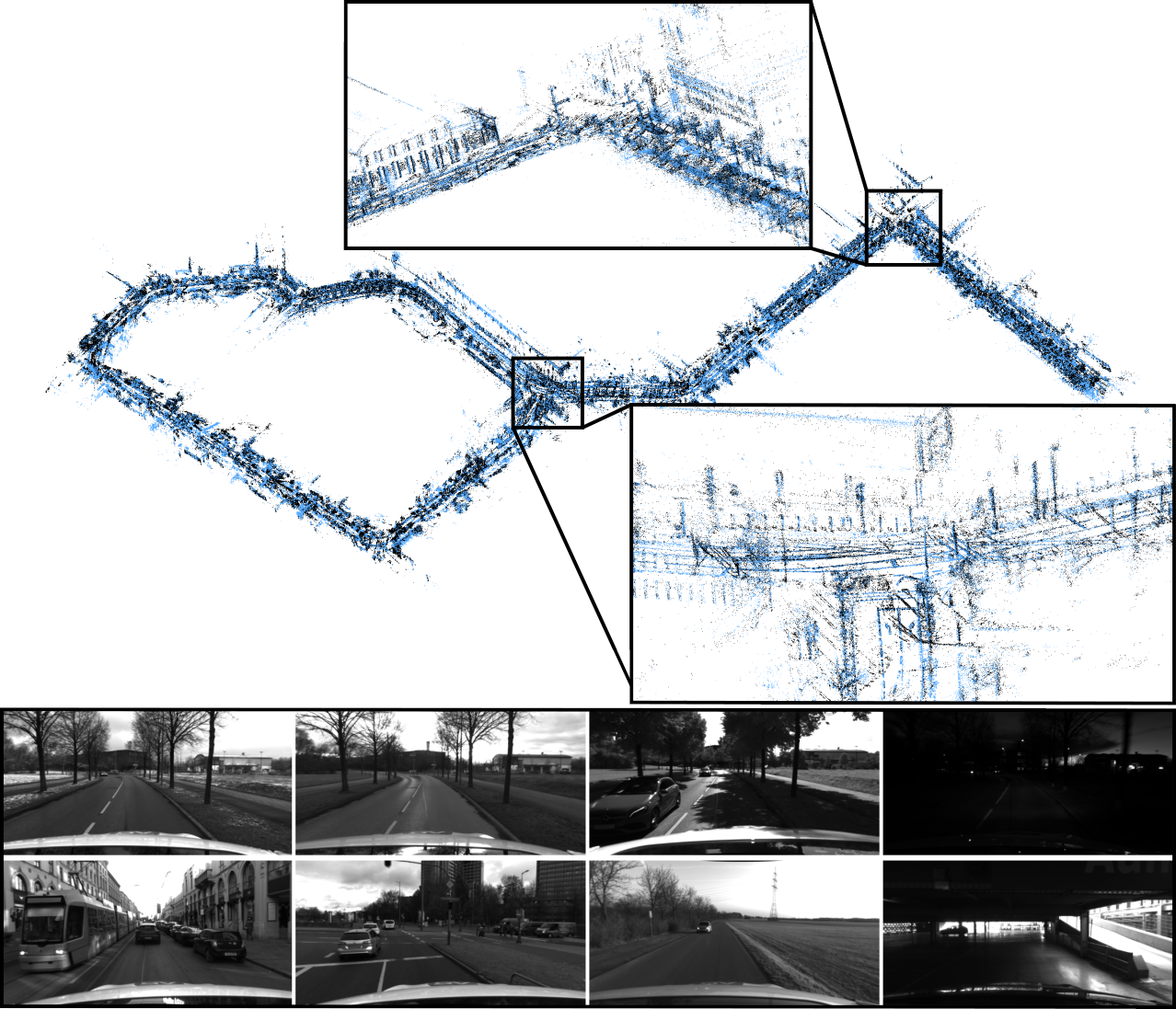}
    \caption{\textbf{Dataset overview.} Top: overlaid maps recorded at different times and environmental conditions. The points from the reference map (black) align well with the points from the query map (blue), indicating that the reference poses are indeed accurate. Bottom: sample images demonstrating the diversity of our dataset. The first row shows a collection from the same scene across different weather and lighting conditions: snowy, overcast, sunny, and night. The second row depicts the variety of scenarios within the dataset: inner city, suburban, countryside, and a parking garage.}
    \label{fig:teaser}
\end{figure}

To accommodate this demand, we present in this paper a versatile cross-season and multi-weather dataset on a large-scale focusing on long-term localization for autonomous driving. By traversing the same stretch under different conditions and over a long-term time horizon, we capture variety in illumination and weather as well as in the appearance of the scenes. Figure~\ref{fig:teaser} visualizes two overlaid 3D maps recorded at different times as well as sample images of the dataset.

In detail this work adds the following contributions to the state-of-the-art:

\begin{itemize}
  \item A cross-season/multi-weather dataset for long-term visual \ac{slam} in automotive applications containing more than \SI{350}{km} of recordings.
  \item Sequences covering nine different kinds of environments ranging from multi-level parking garage over urban (including tunnels) to countryside and highway.
  \item Global \ac{6dof} reference poses with up-to centimeter accuracy obtained from the fusion of direct stereo \ac{vio} with RTK-GNSS.
  \item Accurate cross-seasonal pixel-wise correspondences to train dense feature representations.
\end{itemize}

%% file: sections/related_work.tex
\section{Related Work}\label{sec:related_work}

There exists a variety of benchmarks and datasets focusing on \ac{vo} and \ac{slam} for \ac{ad}. Here, we divide these datasets into the ones which focus only on the task of \ac{vo} as well as those covering different weather conditions and therefore aiming towards long-term \ac{slam}.

\subsection{Visual Odometry}
The most popular benchmark for \ac{ad} certainly is KITTI~\cite{Geiger2013IJRR}. This multi-sensor dataset covers a wide range of tasks including not only \ac{vo}, but also 3D object detection, and tracking, scene flow estimation as well as semantic scene understanding. The dataset contains diverse scenarios ranging from urban over countryside to highway. Nevertheless, all scenarios are only recorded once and under similar weather conditions. Ground truth is obtained based on a high-end \ac{ins}.

Another dataset containing LiDAR, \ac{imu}, and image data at a large-scale is the M\'{a}laga Urban dataset~\cite{Blanco2014malaga}. However, in contrast to KITTI, no accurate \ac{6dof} ground truth is provided and therefore it does not allow for a quantitative evaluation based on this dataset.

Other popular datasets for the evaluation of \ac{vo} and \ac{vio} algorithms not related to \ac{ad} include~\cite{sturm12iros} (handheld RGB-D),~\cite{burri2016euroc} (UAV stereo-inertial), \cite{engel2016monodataset} (handheld mono), and \cite{schubert2018vidataset} (handheld stereo-inertial).

\subsection{Long-Term SLAM}
More related to our work are datasets containing multiple traversals of the same environment over a long period of time. With respect to \ac{slam} for \ac{ad} the Oxford RobotCar Dataset~\cite{maddern20171} represents a kind of pioneer work. This dataset consists of large-scale sequences recorded multiple times for the same environment over a period of one year. Hence, it covers large variations in the appearance and structure of the scene. However, the diversity of the scenarios is only limited to an urban environment. Also, the ground truth provided for the dataset is not accurate up-to centimeter-level and therefore, requires additional manual effort to establish accurate cross-sequence correspondences.

The work by~\cite{SattlerCVPR2018} represents a kind of extension to~\cite{maddern20171}. This benchmark is based on subsequences from~\cite{maddern20171} as well as other datasets. The ground truth of the RobotCar Seasons~\cite{SattlerCVPR2018} dataset is obtained based on \ac{sfm} and LiDAR point cloud alignment. However, due to inaccurate \acsu{gnss} measurements~\cite{maddern20171}, a globally consistent ground truth up-to centimeter-level can not be guaranteed. Furthermore, this dataset only provides one reference traversal in the overcast condition. In contrast, we provide globally consistent reference models for all traversals covering a wide variety of conditions. Hence, every traversal can be used as a reference model that allows further research, \eg on analyzing suitable reference-query pairs for long-term localization and mapping.

\subsection{Other Datasets}
Examples of further multi-purpose \ac{ad} datasets which also can be used for \ac{vo} are~\cite{Cordts2016city,Wang2017toronto,Huang2018apollo,Caesar2019nuscenes}.

As stated in Section~\ref{sec:introduction}, our proposed dataset differentiates from previous related work in terms of being both large-scale (similar to~\cite{Geiger2013IJRR}) as well as having high variations in appearance and conditions (similar to~\cite{maddern20171}). Furthermore, we are providing accurate reference poses based on the fusion of direct stereo \ac{vio} and RTK-GNSS.

%% file: sections/system_overview.tex
\section{System Overview}\label{sec:overview}

This section presents the sensor setup which is used for data recording (Section~\ref{sec:sensor_setup}). Furthermore, we describe the calibration of the entire sensor suite (Section~\ref{sec:calibration}) as well as our approach to obtain up-to centimeter-accurate global \ac{6dof} reference poses (Section~\ref{sec:ground_truth}).

\subsection{Sensor Setup}\label{sec:sensor_setup}
\begin{figure}[t]
\centering
\subfloat[Test vehicle.]{\includegraphics[width=0.4\linewidth]{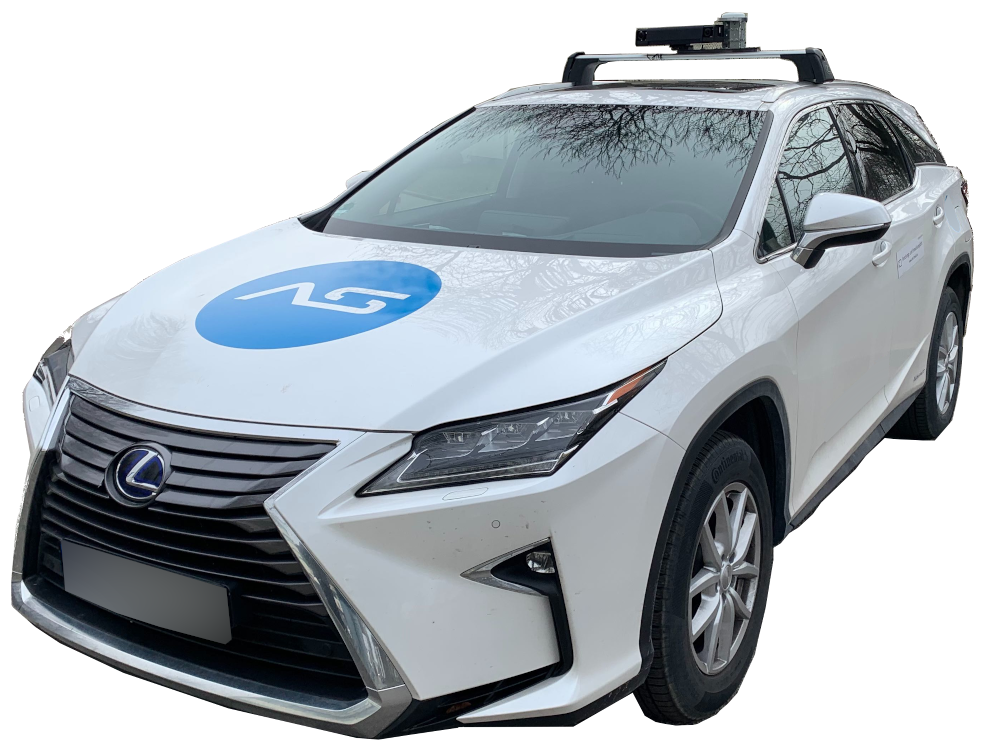}}\hspace{5mm}
\subfloat[Sensor system.]{\includegraphics[width=0.4\linewidth]{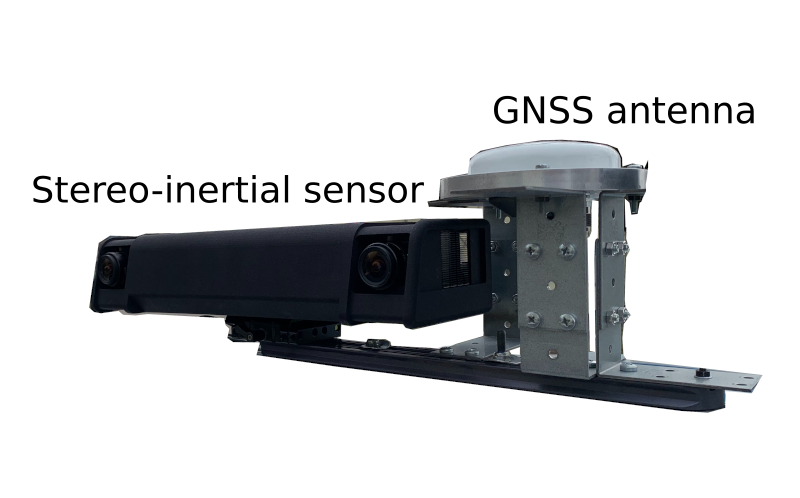}}
\caption{\textbf{Recording setup.} Test vehicle and sensor system used for dataset recording. The sensor system consists of a custom stereo-inertial sensor with a stereo baseline of \SI{30}{cm} and a high-end RTK-GNSS receiver from Septentrio.}
\label{fig:test_vehicle}
\end{figure}

The hardware setup consists of a custom stereo-inertial sensor for \ac{6dof} pose estimation as well as a high-end RTK-GNSS receiver for global positioning and global pose refinement. Figure~\ref{fig:test_vehicle} shows our test vehicle equipped with the sensor system used for data recording.

\subsubsection{Stereo-Inertial Sensor.}
The core of the sensor system is our custom stereo-inertial sensor. This sensor consists of a pair of monochrome industrial-grade global shutter cameras (Basler acA2040-35gm) and lenses with a fixed focal length of $f=\SI{3.5}{mm}$ (Stemmer Imaging CVO GMTHR23514MCN). The cameras are mounted on a highly-rigid aluminum rail with a stereo baseline of \SI{30}{cm}. On the same rail, an \ac{imu} (Analog Devices ADIS16465) is mounted. All sensors, cameras, and \ac{imu} are triggered over an external clock generated by an \ac{fpga}. Here, the trigger accounts for exposure compensations, meaning that the time between the centers of the exposure interval for two consecutive images is always kept constant (1/[frame rate]) independent of the exposure time itself.

Furthermore, based on the \ac{fpga}, the \ac{imu} is properly synchronized with the cameras. In the dataset, we record stereo sequences with a frame rate of \SI{30}{fps}. We perform pixel binning with a factor of two and crop the image to a resolution of $800 \times 400$. This results in a field of view of approximately \ang{77} horizontally and \ang{43} vertically. The \ac{imu} is recorded at a frequency of \SI{2000}{Hz}. During recording, we run our custom auto-exposure algorithm, which guarantees equal exposure times for all stereo image pairs as well as a smooth exposure transition in highly dynamic lighting conditions, as it is required for visual \ac{slam}. We provide those exposure times for each frame. 

\subsubsection{GNSS Receiver.}
For global positioning and to compensate drift in the \ac{vio} system we utilize an RTK-GNSS receiver (mosaic-X5) from Septentrio in combination with an Antcom Active G8 \ac{gnss} antenna. The \ac{gnss} receiver provides a horizontal position accuracy of up-to \SI{6}{mm} by utilizing RTK corrections. While the high-end \ac{gnss} receiver is used for accurate positioning, we use a second receiver connected to the time-synchronization \ac{fpga} to achieve synchronization between the \ac{gnss} receiver and the stereo-inertial sensor.

\subsection{Calibration}\label{sec:calibration}

\subsubsection{Aperture and Focus Adjustment.}
The lenses used in the stereo-system have both adjustable aperture and focus. Therefore, before performing the geometric calibration of all sensors, we manually adjust both cameras for a matching average brightness and a minimum focus blur~\cite{Hu2006ICIP}, across a structured planar target in \SI{10}{m} distance.

\subsubsection{Stereo Camera and IMU.}
For the intrinsic and extrinsic calibration of the stereo cameras as well as the extrinsic calibration and time-synchronization of the \ac{imu}, we use a slightly customized version of \emph{Kalibr}\footnote{\url{https://github.com/ethz-asl/kalibr}}~\cite{rehder2016extending}. The stereo cameras are modeled using the Kannala-Brandt model~\cite{Kannala2006}, which is a generic camera model consisting of in total eight parameters. To guarantee an accurate calibration over a long-term period, we perform a feature-based epipolar-line consistency check for each sequence recorded in the dataset and re-calibrate before a recording session if necessary.

\subsubsection{GNSS Antenna.}
Since the \ac{gnss} antenna does not have any orientation but has an isotropic reception pattern, only the 3D translation vector between one of the cameras and the antenna within the camera frame has to be known. This vector was measured manually for our sensor setup.

\subsection{Ground Truth Generation}\label{sec:ground_truth}
Reference poses (\ie ground truth) for \ac{vo} and \ac{slam} should provide high accuracy in both local relative \ac{6dof} transformations and global positioning. To fulfill the first requirement, we extend the state-of-the-art stereo direct sparse \ac{vo}~\cite{wang2017stereoDSO} by integrating \ac{imu} measurements~\cite{von2018direct}, achieving a stereo-inertial \ac{slam} system offering average tracking drift around \SI{0.6}{\percent} of the traveled distance. To fulfill the second requirement, the poses estimated by our stereo-inertial system are integrated into a global pose graph, each with an additional constraint from the corresponding RTK-GNSS measurement. Our adopted RTK-GNSS system can provide global positioning with up-to centimeter accuracy. The pose graph is optimized globally using the Gauss-Newton method, ending up with \ac{6dof} camera poses with superior accuracy both locally and globally. For the optimization, we make use of the g2o library~\cite{kummerle2011g}.

One crucial aspect for the dataset is that the reference poses which we provide are actually accurate enough, even though some of the recorded sequences partially contain challenging conditions in \ac{gnss}-denied environments. Despite the fact that the stereo-inertial sensor system has an average drift around \SI{0.6}{\percent}, this cannot be guaranteed for all cases. Hence, for the reference poses in our dataset, we report whether a pose can be considered to be reliable by measuring the distance to the corresponding RTK-GNSS measurement. Only RTK-GNSS measurements with a reported standard deviation of less than \SI{0.01}{m} are considered as accurate. For all poses, without corresponding RTK-GNSS measurement we do not guarantee a certain accuracy. Nevertheless, due to the highly accurate stereo-inertial odometry system, these poses still can be considered to be accurate in most cases even in \ac{gnss}-denied environments, \eg tunnels or areas with tall buildings.

%% file: sections/scenarios.tex
\section{Scenarios}\label{sec:scenarios}

This section describes the different scenarios we have collected for the dataset. The scenarios involve different sequences -- ranging from urban driving to parking garage and rural areas. We provide complex trajectories, which include partially overlapping routes, and multiple loops within a sequence. For each scenario, we have collected multiple traversals covering a large range of variation in environmental appearance and structure due to weather, illumination, dynamic objects, and seasonal effects. In total, our dataset consists of nine different scenarios, \ie industrial area, highway, local neighborhood, ring road, countryside, suburban, inner city, monumental site, and multi-level parking garage. 

\begin{figure}[t]
    \centering
    \includegraphics[width=\linewidth]{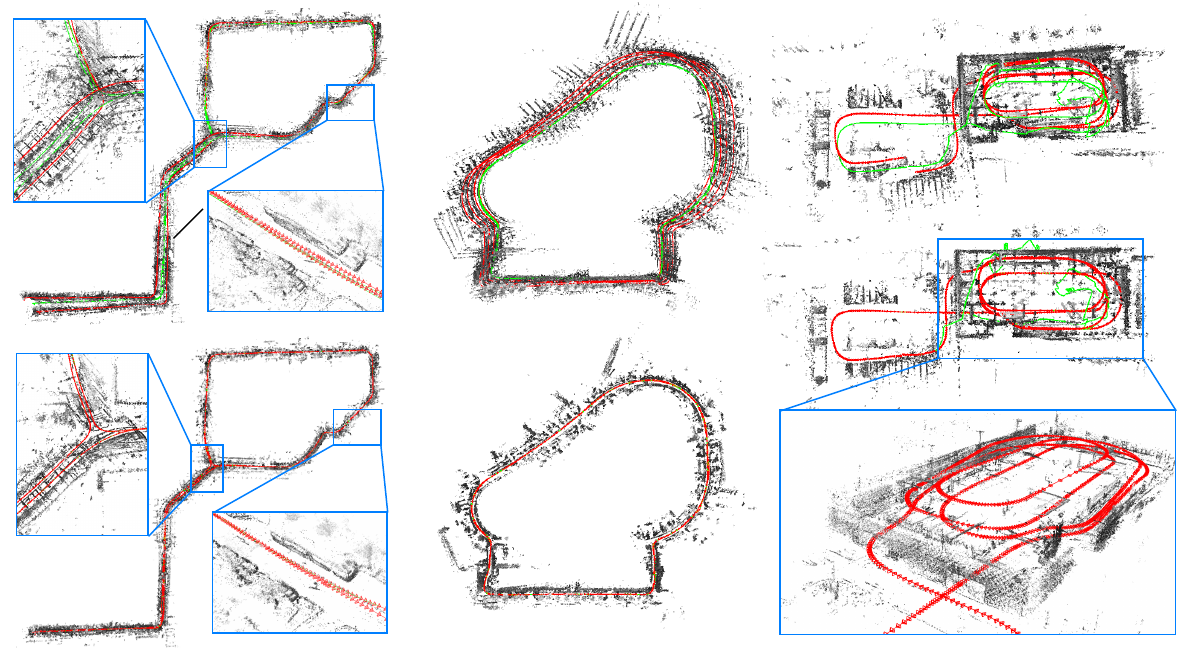}
    \caption{\textbf{3D models of different scenarios contained in the dataset}. The figure shows a loop around an industrial area (left), multiple loops around an area with high buildings (middle), and a stretch recorded in a multi-level parking garage (right). The green lines encode the \ac{gnss} trajectories, and the red lines encode the \ac{vio} trajectories. Top: shows the trajectories before the fusion using pose graph optimization. Bottom: shows the result after the pose graph optimization. Note that after the pose graph optimization the reference trajectory is well aligned.}
    \label{fig:3d_models}
\end{figure}

\begin{figure}[t]
    \centering
    \includegraphics[width=\linewidth]{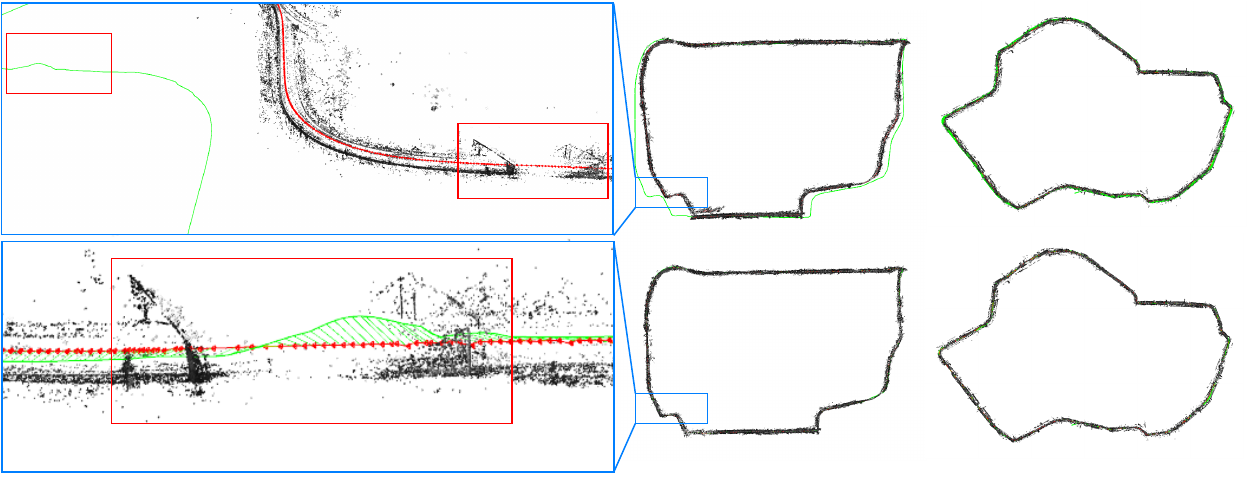}
    \caption{\textbf{Reference poses validation.} This figure shows two additional 3D models of the scenarios collected. Note that these two sequences are quite large (more than \SI{10}{km} and \SI{6}{km}, respectively). Top: before the fusion using pose graph optimization. Bottom: results after optimization. The green lines encode the \ac{gnss} trajectories,  the red lines show the \ac{vio} trajectories (before fusion) and the fused trajectories (after fusion). The left part of the figure shows a zoomed-in view of a tunnel, where the \ac{gnss} signal becomes very noisy as highlighted in the red boxes. Besides, due to the large size of the sequence, the accumulated tracking error leads to a significant deviation of the \ac{vio} trajectory from the \ac{gnss} recordings. Our pose graph optimization, by depending globally on \ac{gnss} positions and locally on \ac{vio} relative poses, successfully eliminates global \ac{vio} drifts and local \ac{gnss} positioning flaws.}
    \label{fig:3d_models_supplementary}
\end{figure}

We provide reference poses and 3D models generated by our ground truth generation pipeline (\cf~Figure~\ref{fig:3d_models}) along with the corresponding raw image frames and raw \ac{imu} measurements. Figure~\ref{fig:3d_models_supplementary} shows another example of the optimized trajectory, which depicts the accuracy of the provided reference poses.

The dataset will challenge current approaches on long-term localization and mapping since it contains data from various seasons and weather conditions as well as from different times of the day as shown in the bottom part of Figure~\ref{fig:teaser}. 

\subsection{Ground Truth Validation}\label{sec:ground_truth_validation}

\begin{figure}[t]
    \centering
    \includegraphics[width=\linewidth]{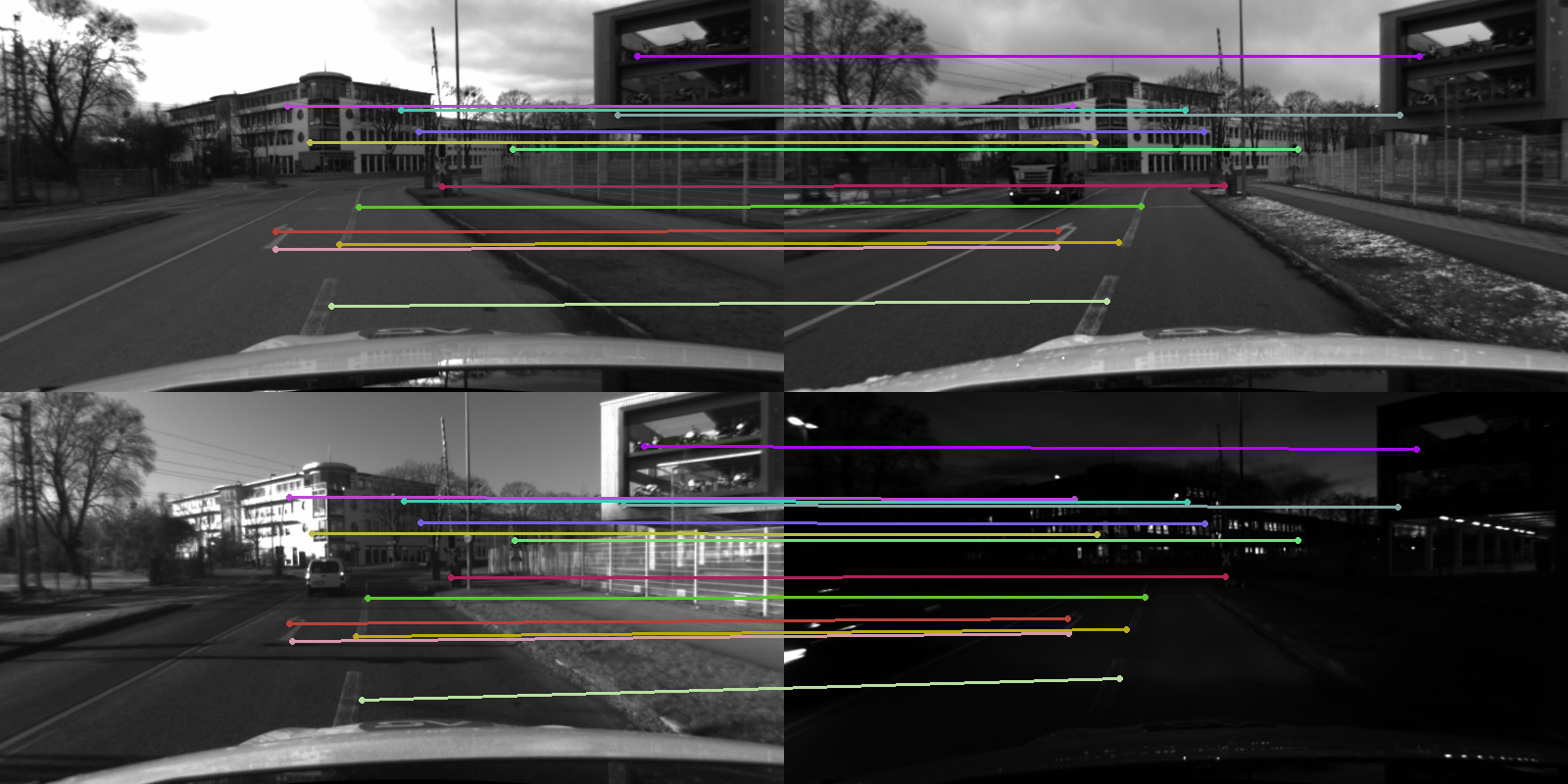}
    \caption{\textbf{Accurate pixel-wise correspondences, making cross-seasonal training possible.} Qualitative assessment of the accuracy of our data collection and geometric reconstruction method for a sample of four different conditions (from top left in clockwise order: \emph{overcast, snowy, night, sunny}) across the same scene. Each same colored point in the four images corresponds to the same geometric point in the world. The cameras corresponding to these images have different poses in the global frame of reference. Please note that the points are not matched but rather a result of our accurate reference poses and geometric reconstruction. This way we are capable of obtaining sub-pixel level accuracy. On average we get more than 1000 of those correspondences per image pair.}
    \label{fig:corres_pts_img}
\end{figure}

The top part of Figure~\ref{fig:teaser} shows two overlaid point clouds from different runs across the same scene. Note that despite the weather and seasonal differences the point clouds align very well. This shows that our reference poses are indeed very accurate. Furthermore, a qualitative assessment of the point-to-point correspondences is shown in Figure~\ref{fig:corres_pts_img}. The figure shows a subset of very accurate pixel-wise correspondences across different seasons (\emph{autumn}/\emph{winter}) in the top and different illumination conditions (\emph{sunny}/\emph{night}) in the bottom. These point-to-point correspondences are a result of our up-to centimeter-accurate global reference poses and are obtained in a completely self-supervised manner. This makes them suitable as training pairs for learning-based algorithms. Recently, there has been an increasing demand for pixel-wise cross-season correspondences which are needed to learn dense feature descriptors~\cite{spencer2020same,Dusmanu2019CVPR,r2d2}. However, there is still a lack of datasets to satisfy this demand. The KITTI~\cite{Geiger2013IJRR} dataset does not provide cross-seasons data. The Oxford RobotCar Dataset~\cite{maddern20171} provides cross-seasons data, however, since the ground truth is not accurate enough, the paper does not recommend benchmarking localization and mapping approaches.

Recently, RobotCar Seasons~\cite{SattlerCVPR2018} was proposed to overcome the inaccuracy of the provided ground truth. However, similar to the authors of~\cite{spencer2020same}, we found that it is still challenging to obtain accurate cross-seasonal pixel-wise matches due to pose inconsistencies. Furthermore, this dataset only provides images captured from three synchronized cameras mounted on a car, pointing to the rear-left, rear, and rear-right, respectively. Moreover, the size of the dataset is quite small and a significant portion of it suffers from strong motion blur and low image quality.

To the best of our knowledge, our dataset is the first that exhibits accurate cross-season reference poses for the \ac{ad} domain.

%% file: sections/tasks.tex
\section{Tasks}\label{sec:tasks}
This section describes the different tasks of the dataset. The provided globally consistent \ac{6dof} reference poses for diverse conditions will be valuable to develop and improve the state-of-the-art for different \ac{slam} related tasks. Here the major tasks are robust \ac{vo}, global place recognition, and map-based re-localization tracking.

In the following, we will present the different subtasks for our dataset. 

\subsection{Visual Odometry in Different Weather Conditions}\label{subsec:odometry_benchmark}
\ac{vo} aims to accurately estimate the \ac{6dof} pose for every frame relative to a starting position. To benchmark the task of \ac{vo} there already exist various datasets~\cite{Geiger2012CVPR,sturm12iros,engel2016monodataset}. All of these existing datasets consist of sequences recorded at rather homogeneous conditions (indoors, or sunny/overcast outdoor conditions). However, especially methods developed for \ac{ad} use cases must perform robustly under almost any condition. We believe that the proposed dataset will contribute to improving the performance of \ac{vo} under diverse weather and lighting conditions in an automotive environment. Therefore, instead of replacing existing benchmarks and datasets, we aim to provide an extension that is more focusing on challenging conditions in \ac{ad}. As we provide frame-wise accurate poses for large portions of the sequences, metrics well known from other benchmarks like \ac{ate} or \ac{rpe}~\cite{Geiger2012CVPR,sturm12iros} are also applicable to our data.

\subsection{Global Place Recognition}\label{subsec:global_image_retrieval}
Global place recognition refers to the task of retrieving the most similar database image given a query image~\cite{lowry2015visual}. In order to improve the searching efficiency and the robustness against different weather conditions, tremendous progress on global descriptors~\cite{jegou2010aggregating,arandjelovic2013all,angeli2008fast,galvez2012bags} has been seen. For the re-localization pipeline, visual place recognition serves as the initialization step to the downstream local pose refinement by providing the most similar database images as well as the corresponding global poses. Due to the advent of deep neural networks~\cite{simonyan2014very,krizhevsky2012imagenet,he2016deep,szegedy2015going}, methods aggregating deep image features are proposed and have shown advantages over classical methods~\cite{arandjelovic2016netvlad,gordo2016deep,radenovic2018fine,tolias2015particular}.

\begin{figure}[t]
    \centering
    \includegraphics[width=\linewidth]{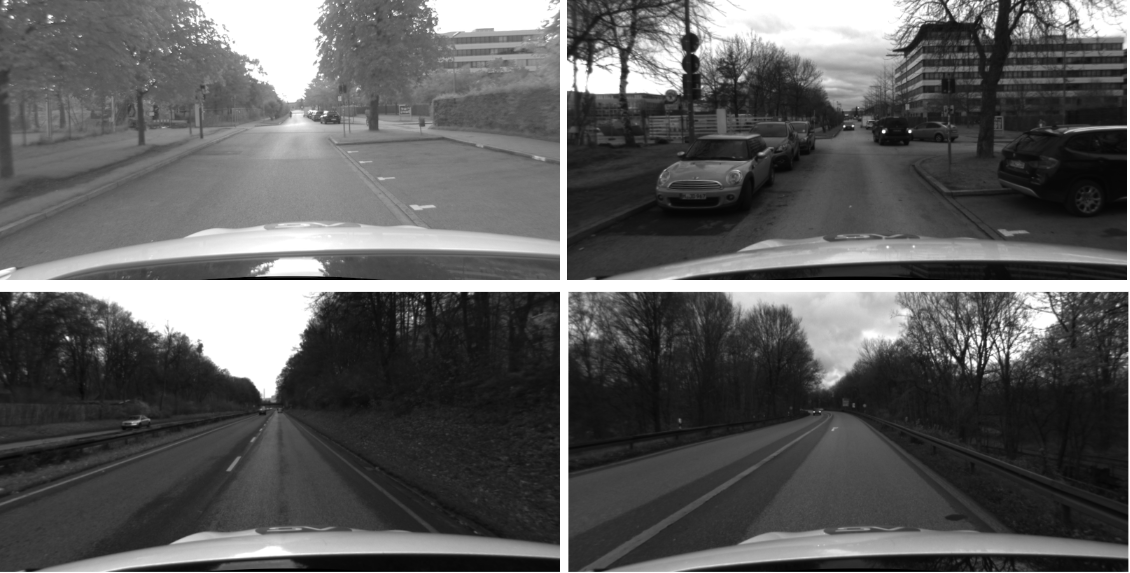}
    \caption{\textbf{Challenging scenes for global place recognition.} Top: two pictures share the same location with different appearances. Bottom: two pictures have similar appearance but are taken at different locations.}
    \label{fig:vpr}
\end{figure}

The proposed dataset is challenging for global place recognition since it contains not only cross-season images that have different appearances but share a similar geographical location but also the intra-season images which share similar appearances but with different locations. Figure~\ref{fig:vpr} depicts example pairs of these scenarios. We suggest to follow the standard metric widely used for global place recognition~\cite{arandjelovic2016netvlad,arandjelovic2013all,sattler2012image,gordo2016deep}. 

\subsection{Map-Based Re-Localization Tracking}\label{subsec:reloc_tracking}
Map-based re-localization tracking~\cite{gn-net-2020} refers to the task of locally refining the \ac{6dof} pose between reference images from a pre-built reference map and images from a query sequence. In contrast to wide-baseline stereo matching, for re-localization tracking, it is also possible to utilize the sequential information of the sequence. This allows us to estimate depth values by running a standard \ac{vo} method. Those depth estimates can then be used to improve the tracking of the individual re-localization candidates.

In this task we assume to know the mapping between reference and query samples. This allows us to evaluate the performance of local feature descriptor methods in isolation. In practice, this mapping can be found using image retrieval techniques like NetVLAD~\cite{arandjelovic2016netvlad} as described in Section~\ref{subsec:global_image_retrieval} or by aligning the point clouds from the reference and query sequences~\cite{SattlerCVPR2018}, respectively. 

Accurately re-localizing in a pre-built map is a challenging problem, especially if the visual appearance of the query sequence significantly differs from the base map. This makes it extremely difficult especially for vision-based systems since the localization accuracy is often limited by the discriminative power of feature descriptors. Our proposed dataset allows us to evaluate re-localization tracking across multiple types of weather conditions and diverse scenes, ranging from urban to countryside driving. Furthermore, our up to centimeter-accurate ground truth allows us to create diverse and challenging re-localization tracking candidates with an increased level of difficulty. By being able to precisely changing the re-localization distances and the camera orientation between the reference and query samples, we can generate more challenging scenarios. This allows us to determine the limitations and robustness of current state-of-the-art methods.

%% file: sections/conclusion.tex
\section{Conclusion}
We have presented a cross-season dataset for the purpose of multi-weather \ac{slam}, global visual localization, and local map-based re-localization tracking for \ac{ad} applications. Compared to other datasets, like KITTI~\cite{Geiger2013IJRR} or Oxford RobotCar~\cite{maddern20171}, the presented dataset provides diversity in both multiplicities of scenarios and environmental conditions. Furthermore, based on the fusion of direct stereo \ac{vio} and RTK-GNSS we are able to provide up-to centimeter-accurate reference poses as well as highly accurate cross-sequence correspondences. One drawback of the dataset is that the accuracy of the reference poses can only be guaranteed in environments with good \ac{gnss} receptions. However, due to the low drift of the stereo \ac{vio} system, the obtained reference poses are also very accurate in \ac{gnss}-denied environments, \eg tunnels, garages, or urban canyons.

We believe that this dataset will help the research community to further understand the limitations and challenges of long-term visual \ac{slam} in changing conditions and environments and will contribute to advance the state-of-the-art. To the best of our knowledge, ours is the first large-scale dataset for \ac{ad} providing cross-seasonal accurate pixel-wise correspondences for diverse scenarios. This will help to vastly increase robustness against environmental changes for deep learning methods. The dataset is made publicly available to facilitate further research.